\documentclass[a4paper]{article}

\usepackage{INTERSPEECH2019}
\usepackage{multirow}
\usepackage{colortbl}
\usepackage{url}
\usepackage{bbm}

\title{End-to-End Speech Translation with Knowledge Distillation}
\name{Yuchen Liu$^{1,2}$, Hao Xiong$^4$, Zhongjun He$^4$,  Jiajun Zhang$^{1,2}$, Hua Wu$^4$, Haifeng Wang$^4$ and Chengqing Zong$^{1,2,3}$}
\address{
	$^1$NLPR, Institute of Automation, Chinese Academy of Sciences,  Beijing, China\\
	$^2$University of Chinese Academy of Sciences, China\\
	$^3$CAS Center for Excellence in Brain Science and Intelligence Technology, Beijing, China\\
    $^4$Baidu Inc. No. 10, Shangdi 10th Street, Beijing, 100085, China
}
\email{\{yuchen.liu, jjzhang, cqzong\}@nlpr.ia.ac.cn, \\
	\{xionghao05, hezhongjun, wu\_hua, wanghaifeng\}@baidu.com}

\begin{document}
	
	\maketitle
	\begin{abstract}
		End-to-end speech translation (ST), which directly translates from source language speech into target language text, has attracted intensive attentions in recent years. Compared to conventional pipepine systems, end-to-end ST models have advantages of lower latency, smaller model size and less error propagation.
		However, the combination of speech recognition and text translation in one model is more difficult than each of these two tasks.
		In this paper, we propose a \textit{knowledge distillation} approach to improve ST model by transferring the knowledge from text translation model.
		Specifically, we first train a text translation model, regarded as a teacher model, and then ST model is trained to learn output probabilities from teacher model through knowledge distillation.
		Experiments on English-French Augmented LibriSpeech and English-Chinese TED corpus show that end-to-end ST is possible to implement on both similar and dissimilar language pairs. In addition, with the instruction of teacher model, end-to-end ST model can gain significant improvements by over 3.5 BLEU points.
		
	\end{abstract}
	
	\section{Introduction}
	Conventional speech translation system is a pipeline of two main components: an automatic speech recognition (ASR) model which provides transcripts of source language utterances, and a text machine translation (MT) model  which translates the transcripts to target language \cite{Bahdanau:2015,chan2016listen,Wu:2016,chiu2017state,vaswani2017attention}. This pipeline system usually suffers from time delay, parameter redundancy and error accumulation.  
	In contrast, end-to-end ST, based on an encoder-decoder architecture with attention mechanism, is more compact and efficient. It can directly generate translations from raw audio and jointly optimize parameters on the final goal. Therefore, this model has become a new trend in speech translation research studies \cite{anastasopoulos2016unsupervised,duong2016attentional,berard2016listen,weiss2017sequence,berard2018end,bansal2018pre}.
	
	However, despite appealing advantages of end-to-end ST model, its performance is generally inferior. One of the important reasons is due to extremely scarce data which includes speech in source language paired with text in target language. Previous studies resort pretraining or multi-task learning approaches to improve the translation quality. They either pretrain ASR task on high-resource data \cite{bansal2018pre}, or use multi-task learning to train ST model with ASR or MT model simultaneously \cite{weiss2017sequence,berard2018end}. Nevertheless, they only gain limited improvements and do not take full advantage of text data.  We notice that the performance between end-to-end ST and MT model exists a huge gap, thus how to utilize MT model to help instruct end-to-end ST model is of great significance.
	
	It is a challenge to train an end-to-end ST model directly from speech signal without text guidance while achieving comparable performance as text translation model. Given that text translation models are superior to ST model, we consider ST model can be improved by leveraging \textit{knowledge distillation}. In knowledge distillation, there is usually a big teacher model with a small student model. It has been shown that the output probabilities of teacher model are smooth, which are easier for student model to learn from than ground-truth text \cite{hinton2015distilling}. Thus, a student model can be taught by imitating the behaviour of teacher model, such as output probabilities \cite{hinton2015distilling, freitag2017ensemble}, hidden representation \cite{yim2017gift,romero2014fitnets}, or generated sequence \cite{kim2016sequence}, and alleviate the performance gap between itself and the teacher model.
	
	In this paper, we present a method based on knowledge distillation for end-to-end ST model to learn knowledge from text translation model. We first train a text translation model on parallel text data (regarded as teacher) and then an end-to-end ST model (regarded as student) is trained by learning from ground-truth translations and the outputs of teacher model simultaneously. 
	Experiments conducted on 100h English-French Augmented LibriSpeech corpus and 542h English-Chinese TED corpus show that it is possible to train a compact end-to-end speech translation model on both similar and dissimilar language pairs. With the instruction of teacher model, end-to-end ST model can gain significant improvements,  approaching to the traditional pipeline system.
	
	\section{Related Work}
	End-to-end model has already become a dominant paradigm in machine translation task, which adopts an encoder-decoder architecture and generates target words from left to right at each step \cite{Bahdanau:2015,Wu:2016,vaswani2017attention}. This model has also achieved promising results in ASR fields \cite{chan2016listen,chiu2017state,bahdanau2016end}. 
	Recent works purpose a further attempt to combine these two tasks together by building an end-to-end speech-to-text translation without the use of source language translation during learning or decoding. 
	
	Anastasopoulos et al. \cite{anastasopoulos2016unsupervised} use $k$-means clustering to cluster repeated audio patterns and automatically align spoken words with their translations. Duong et al. \cite{duong2016attentional} focus on the alignment between speech and  translated phrase but not to directly predict the final translations. 
	B\'{e}rard et al. \cite{berard2016listen} give the first proof of the potential for end-to-end speech-to-text translation without using source language. They further conduct experimetns on a larger English-to-French dataset and pre-train encoder and decoder which improves performance \cite{berard2018end}.
	Weiss et al. \cite{weiss2017sequence} also use multi-task learning and show that end-to-end model can outperform a cascade of independently trained pipeline system on Fisher Callhome Spanish-English speech translation task. 
	Bansal et al. \cite{bansal2018pre} find pretraining encoder on higher-resource language ASR training data can achieve gains in low-resource speech translation system. However, these work mainly focus on pretraining acoustic encoder and do not take full advantage of text data.
	
	Knowledge distillation is first adopted to apply for model compression, the main idea of which is to train a smaller student model to mimic a larger teacher model, by minimizing the loss between the teacher and student predictions. 
	It has soon been applied to a variety of tasks, like image classification \cite{hinton2015distilling,li2017learning,yang2018knowledge,anil2018large}, speech recognition \cite{hinton2015distilling} and natural language processing \cite{freitag2017ensemble,kim2016sequence,tan2019multilingual}.
	The teacher and student model in conventional knowledge distillation usually handle the same task, while in our method the teacher model and student model have different input modalities where teacher uses text as input and student uses speech.
	
	\section{Models}
	In this paper, we apply end-to-end models with the same architecture for all three tasks (ASR, ST and MT). The model architecture is similar with \textit{Transformer} \cite{vaswani2017attention}, which is the state-of-art model in MT task. Recently, this model also begins to be used in ASR task, showing a decent performance \cite{dong2018speech,zhou2018syllable}. In this section, we first describe the core architecture of \textit{Transformer} and then show how this model is applied to ASR/ST and MT task.
	
	\subsection{Core Module of Transformer}
	\textit{Transformer} is an encoder-decoder architecture which entirely relies on self-attention mechanism including scaled dot-product attention and multi-head attention. 
	It consists of $N$ stacked encoder and decoder layers. Each encoder layer has two blocks, which is a self-attention block followed by a feed-forward block. Decoder layer has the same architecture with encoder layer except an extra encoder-decoder attention block to perform attention over the output of the top encoder layer. Residual connection and layer normalization are employed around each block. In addition, the self-attention block in the decoder is modified with mask to prevent present positions attending to future positions during training.
	
	To be detailed, multi-head attention technique is applied in self-attention and encoder-decoder attention blocks  to obtain information from different representation subspaces at different positions. 
	Each head is corresponding to a scaled dot-product attention, which operates on query Q, key K and value V:
	\begin{equation}
	\mathrm{Attention(\mathbf{Q},\mathbf{K},\mathbf{V})}=\mathrm{softmax}(\frac{\mathbf{Q}\mathbf{K}^T}{\sqrt{d_k}})\mathbf{V} \label{attention}
	\end{equation}
	where $d_k$ is the dimension of the key.
    Then the output values are concatenated, 
	\begin{align}
	\mathrm{MultiHead}(\mathbf{Q},\mathbf{K},\mathbf{V}) &= \mathrm{Concat}(\mathrm{head_1},\cdots,\mathrm{head_h})\mathbf{W^O} \notag  \\
	\mathrm{where} \  \mathrm{head_i} &= \mathrm{Attention}(\mathbf{Q}\mathbf{W}_i^Q, \mathbf{KW}_i^K, \mathbf{VW}_i^V)
	\label{multihead}
	\end{align}
	where the $\mathbf{W}_i^Q\in \mathbb{R}^{d_{model} \times d_q}$, $\mathbf{W}_i^K \in \mathbb{R}^{d_{model} \times d_k}$, $\mathbf{W}_i^V \in \mathbb{R}^{d_{model} \times d_v}$ and $\mathbf{W}^O \in \mathbb{R}^{d_v\times d_{model}}$ are projection matrices that are learned. $d_q=d_k=d_v=d_{model/h}$, $h$ is the number of heads.
	
	Position-wise feed-forward block is composed of two linear transformations with a ReLU activation in between.
	\begin{equation}
	\mathrm{FFN}(x)=max(0,x\mathbf{W}_1+\mathbf{b}_1)\mathbf{W}_2+\mathbf{b}_2
	\end{equation}
	where the weights $\mathbf{W}_1\in  \mathbb{R}^{d_{model} \times d_{ff}}$, $\mathbf{W}_2\in  \mathbb{R}^{d_{ff} \times d_{model}}$ and the biases $\mathbf{b}_1\in \mathbb{R}_{d_{ff}}$, $\mathbf{b}_2\in \mathbb{R}^{d_{model}}$.
	
	
	\begin{figure}[t]
		\centering
		\includegraphics[scale=0.7]{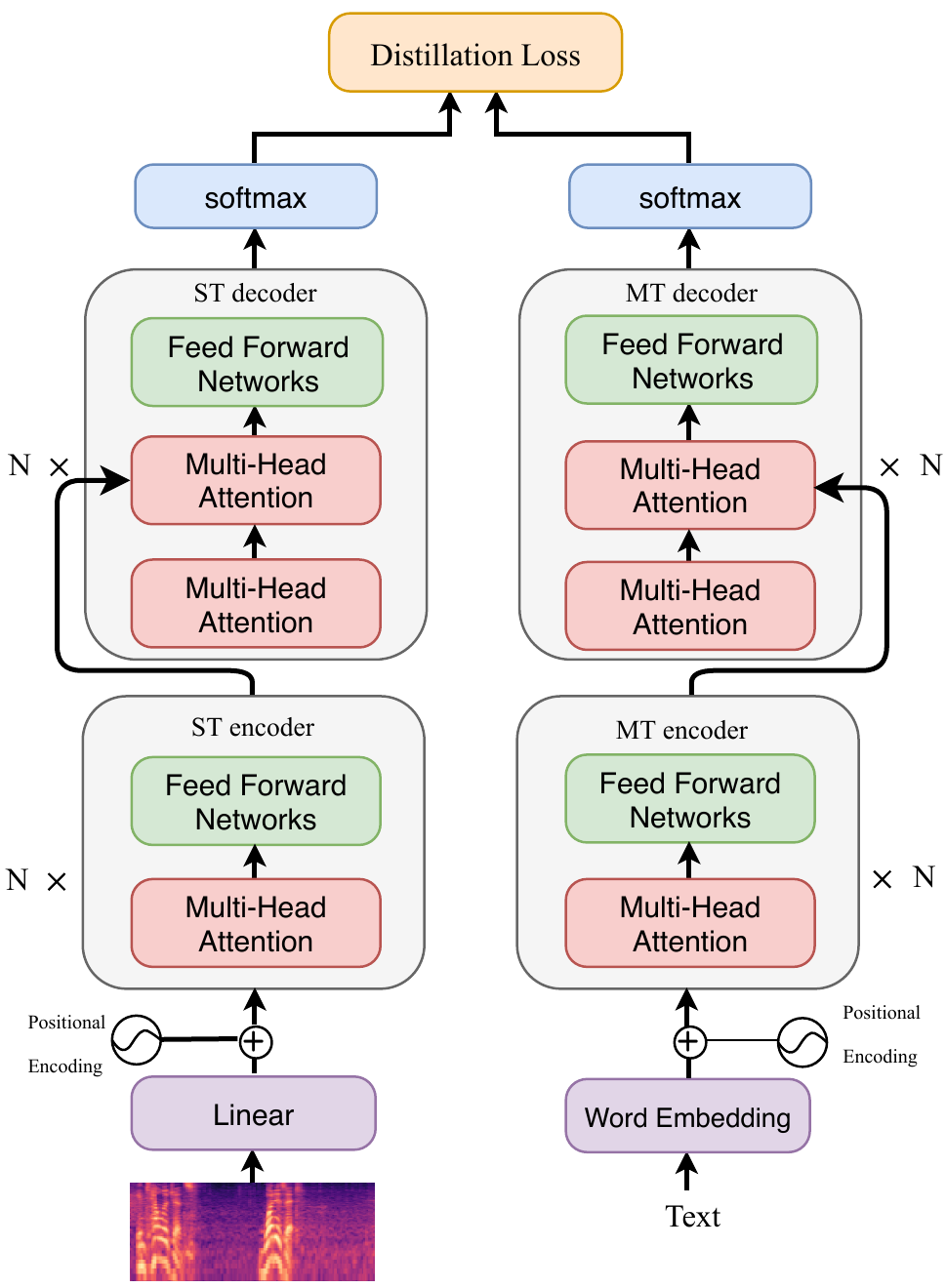}
		\caption{Model architecture of our method. The left part is ST model, regarded as a student model, whose input is speech signal. The right part is MT model, regarded as a teacher model, whose input is the source sentence corresponding to the input of student model. The output of both student model and teacher model is target sentence. The top part is distillation loss, where the student model not only matches the ground-truth, but also the output probabilities of the teacher model.}
		\label{fig:transformer}
	\end{figure}
	
	\subsection{ASR/ST Model}
	The ASR/ST model is shown in the left part of Figure \ref{fig:transformer}, whose input is a series of discrete-time speech signal. We first use log-Mel filterbank to convert raw speech signal into a sequence of acoustic features and then apply mean and variance normalization. 
	To prevent the GPU memory overflow and produce approximate hidden representation length against target length, we apply frame stack and downsample similar to \cite{sak2015fast,kannan2018analysis}. The final acoustic feature sequence is $S=(s_1,s_2,\cdots,s_n)$ with dimension of $d_{filterbank} \times num_{stack}$.  Then the feature sequence is fed into a linear transformation with a normalization layer to map with model dimension $d_{model}$.  In addition, positional encodings are added to the feature sequence in order to enable the model to attend by relative positions. This sequence is finally treated as the input into \textit{Transformer} model.  Other parts are the same with \textit{Transformer} model.  For ASR the input to decoder is source language text, while the input to decoder in ST is target language text. 
	
	
	\subsection{MT Model}
    We also use \textit{Transformer} to train a baseline MT model, as shown in the right part of Figure \ref{fig:transformer}. The difference between MT model and ASR/ST model is the input to the encoder. In MT model, $\mathrm{X}=(x_1,x_2,\cdots,x_n)$ is a sequence of tokens, representing source sentence. We embed the words in sequence \emph{X} into a real continuous space with the dimension of $\mathbb{R}^d_{model}$, which can be fed into a neural network. 
	\subsection{Knowledge Distillation}
	Training an end-to-end ST model is considerably difficult than MT model. The accuracy of the later model is usually much higher than the former. Therefore, we present MT model as a teacher to teach ST model. Here we give a description of the idea of knowledge distillation.
	
    Denote $D={(s,x,y)}$ as the corpus of triple data corresponding to speech signal, transcription in source language and its translation. 
	The log-likelihood loss of ST model can be formulated as follows:
	\begin{equation}
	L_{\mathrm{ST}}(D;\theta)=-\sum_{(s,y)\in D} \mathrm{log} P(y|s;\theta)
	\end{equation}
	\begin{equation}
	\mathrm{log} P(y|s;\theta)=\sum_{t=1}^{N} \sum_{k=1}^{|V|} \mathbbm{1} (y_t=k) \mathrm{log} P(y_t=k|y_{<t},s;\theta)
	\end{equation}
	where $s$ is the acoustic feature sequence of source speech signal, $y$ is the target translated sentence, 
	$N$ is the length of the output sequence, $|V|$ is the vocabulary size of the output language, $y_t$ is the $t$-th output token, $\mathbbm{1}(y_t=k)$ is an indicator function which indicates whether the output token is equal to the ground-truth.
	
    We denote the output distribution of teacher model for token $y_t$ as $Q(y_t|y_{<t},x;\theta_T)$, and $x$ is the source transcribed sentence which corresponds to speech signal $s$. Then the cross entropy between the distributions of teacher and student is:
	\begin{align}
	L_{\mathrm{KD}}(D;\theta,\theta_T) =-\sum_{(x,y)\in D}\sum_{t=1}^{N} \sum_{k=1}^{|V|} &Q(y_t=k|y_{<t},x;\theta_T) \notag  \\ 
	\mathrm{log} & P(y_t=k|y_{<t},x;\theta) 
	\end{align}
	
	In distillation loss, the student not only matches the output of ground-truth, but also the output probabilities of teacher model, which is more smooth and yields smaller variance in gradients \cite{hinton2015distilling}. Then the total loss function is,
	\begin{equation}
	L_{\mathrm{ALL}}(D;\theta;\theta_T)=(1-\lambda)L_{\mathrm{ST}}(D;\theta) + \lambda L_{\mathrm{KD}}(D;\theta,\theta_T)
	\end{equation}
	where $\lambda$ is a hyper-parameter to trade off these two loss terms.
	
	\section{Experiments}
	\subsection{Datasets}
	We conduct experiments on Augmented LibriSpeech which is collected by \cite{kocabiyikoglu2018augmenting} and available for free. This corpus is built by automatically aligning e-books in French with English utterances of LibriSpeech, which contains 236 hours of speech in total. They provide quadruplet: English speech signal, English transcription, French text translations from alignment of e-books and \textit{Google Translate} references. Following \cite{berard2018end}, We only use the 100 hours clean train set for training, with 2 hours development set and 4 hours test set, which corresponds to 47,271, 1071 and 2048 utterances respectively. To be consistent with their settings, we also double the training size by concatenating the aligned references with the Google Translate references.
	
	To verify whether the end-to-end speech translation model can handle on dissimilar language pairs, we build a corpus in English-Chinese direction. The raw data (including video, subtitles and timestamps) are crawled from TED website\footnote{\url{https://www.ted.com}}. For each talk, we build a \textit{wav} audio file extracted from video by ffmpeg\footnote{\url{http://ffmpeg.org}}. We also collect its corresponding transcript and save in \textit{txt} format. 
	We divide each audio file into small segments based on timestamps instead of voice activity detection (VAD), because it eliminates the influence of improper fragments and guarantees each utterance containing complete semantic information, which is important for translation. In the end, we totally get 317,088 utterances ($\sim$542 hours). Development and test sets are split according to the partition in IWSLT. We use dev2010 as development set and tst2015 as test set, which has 835 utterances ($\sim$1.48 hours) and 1,223 utterances ($\sim$2.37 hours) respectively. The remaining data are put into training set. We will release this dataset to public as a benchmark soon.
	
	\subsection{Experimental Setup}
	Our acoustic features are 80-dimensional log-Mel filterbanks extracted with a step size of 10ms and window size of 25ms and extended with mean subtraction and variance normalization. The features are stacked with 3 frames to the left and downsample to a 30ms frame rate. For text data, we lowercase all the texts, tokenize and apply normalize punctuations with the Moses scripts\footnote{\url{https://www.statmt.org/moses/}}. For Augmented LibriSpeech corpus, we apply BPE \cite{sennrich2015neural} on the combination of English and French text to obtain subword units. The number of merge operations in BPE is set to 8K, resulting in a shared vocabulary with 8,159 subwords. For TED English-Chinese, the  merge number is 30K, and vocabulary size are 28,912 and 30,000, respectively. We report case-insensitive BLEU scores \cite{papineni2002bleu} by multi-bleu.pl script for the evaluation of ST and MT tasks and use word error rates (WER) to evaluate ASR task. 
	
	Because the size of Augmented LibriSpeech is relatively small,  we set the hidden size $d_{model}=256$, the filter size in feed-forward layer $d_{ff}=1024$, the head number $h=8$, the residual dropout and attention dropout are 0.1. 
	For TED English-Chinese, we set the hidden size $d_{model}=512$ with the filter size $d_{ff}=2048$.  
	MT model,  as a teacher model,  can use bigger parameters. We use 512 hidden sizes, 2048 filter sizes with 8 heads.The number of encoder layers and decoder layers in above models are all set to 6.
	We train our models with Adam optimizer \cite{kingma2014adam} with $\beta_1=0.9$, $\beta_2=0.98$ and $\epsilon=10^{-9}$ on 2 NVIDIA V100 GPUs. 

	\subsection{Results}
	Table \ref{tab:ASR and MT} shows the results for the ASR and MT tasks on Augmented LibriSpeech. It can be seen that \textit{Transformer} model significantly outperforms in both ASR and MT tasks, with 0.92 WER reduction and 4.1 BLEU scores improvement in beam search compared to \cite{berard2018end}. We contribute it to the superior performance of \textit{Transformer} model which is good at modeling long distance in sequence-to-sequence tasks, especially for MT tasks. Contrary to \cite{berard2018end} which uses characters as output units, we consider subword units can also obtain improvements.
	
	For ST task, we have four settings. The pipeline model uses ASR outputs as MT inputs, where ASR model and MT model are described above. The \textit{end-to-end} model is directly trained on source speech signal paired with target text translations. The \textit{pre-trained} model is identical to \textit{end-to-end} model, but it is initialized with ASR and MT models. 
	Knowledge distillation (KD) is our method which uses MT model as teacher model to instruct \textit{end-to-end} ST model. 
	
	\begin{table}[t]
		\centering
		\caption{ASR and MT results on test set of Augmented LibriSpeech.}
		\begin{tabular}{c|c|c|c}
			\hline
			LibriSpeech & Method & WER($\downarrow$) & BLEU($\uparrow$)  \\
			\hline
			\multirow{2}{*}{B\'{e}rard  \cite{berard2018end}} & greedy & 19.9  &  19.2 \\ 
			& beam search & 17.9 & 18.8 \\
			\hline
			\multirow{2}{*}{Ours} & greedy & 21.46  & 21.35 \\
			& beam search & \textbf{16.98}  & \textbf{22.91} \\
			\hline
		\end{tabular}
		\label{tab:ASR and MT}
	\end{table}
	
	As shown in Table \ref{tab:ST}, all four settings surpass the results in \cite{berard2018end}. 
	Noticing that there exists a huge gap between the performance of the end-to-end ST model and MT model, even if the end-to-end ST model is pretrained, thus we conduct \textit{knowledge distillation} to instruct ST model with MT model. The result shows that this method can bring significant improvement on the BLEU score which increases from 14.30 to 17.02. With the instruction of MT model, the performance gap is alleviated, approaching to the pipeline system. It demonstrates the effectiveness of our method.
	\begin{table}[t]
		\centering
		\caption{ST results on Augmented LibriSpeech test. KD denotes knowledge distillation.}
		\begin{tabular}{c|c|c|c|c}
			\hline
			LibriSpeech & Method & greedy & beam  & ensemble \\
			\hline
			\multirow{3 }{*}{B\'{e}rard \cite{berard2018end}}  & Pipeline & 14.6  & 14.6  &  15.8 \\ 
			\cline{2-5}
			& End-to-end & 12.3 & 12.9 & \multirow{2}{*}{15.5} \\
			& Pre-trained &  12.6 & 13.3  &  \\ 
			\hline
			\multirow{4}{*}{Ours} & Pipeline & 15.75  & 17.85   &  18.4  \\ 
			\cline{2-5}
			& End-to-end & 10.19  & 13.15  & \multirow{3}{*}{ 17.8 } \\
			& Pre-trained & 13.89  & 14.30  &  \\ 
			& KD & \textbf{14.96}  &  \textbf{17.02} &  \\ 
			\hline
		\end{tabular}
		\label{tab:ST}
	\end{table}
	
	We also conduct experiments on English-Chinese to verify our methods. Table \ref{tbl:chinese} presents the results of MT and ST models. Pipeline model combines both the ASR (WER is  18.2\%) and MT models. It is difficult to train end-to-end ST model from random initialization parameters, for the reordering between dissimilar language pairs is difficult to align with frame based speech representations. The end-to-end ST model here is pretrained with ASR. With knowledge distillation, it can obtain significant simprovements, proving the generality of our method. Although end-to-end ST does not outperform pipeline system, it shows the potential to implement a compact end-to-end model even on dissimilar language pairs. 
	\begin{table}[t]
		\centering
		\caption{MT and ST results on TED English-Chinese test.}
		\begin{tabular}{c|c|c|c|c}
			\hline
			TED & MT & Pipeline & End-to-end & KD \\
			\hline
			BLEU &  27.08 & 22.28  & 16.80  & 19.55 \\
			\hline
		\end{tabular}
		\label{tbl:chinese}
	\end{table}
	
	\subsection{Analysis}
	To evaluate the effect of teacher model, we  explore different hyper-parameters $\lambda$ of the distillation loss on Augmented LibriSpeech. With $\lambda$ increasing, ST will pay more attention to the teacher model. When $\lambda$ equals 0, it is the pre-trained end-to-end model; when $\lambda$ is 1, it will ignore ground-truth and only learn from the teacher. As Table \ref{tab:weight} shows, the performance becomes better with the increasing of $\lambda$. End-to-end ST obtains the best performance when it only learns the output distributions of teacher model.
	\begin{table}[t]
		\centering
		\caption{The effect of teacher model weight on ST results.}
		\begin{tabular}{c|c|c|c|c|c|c}
			\hline
			$\lambda$ & 0.0 & 0.2 & 0.4 & 0.6 & 0.8 & 1.0  \\
			\hline
			BLEU & 14.30  & 15.68  &  16.73  & 16.62 &  16.93 & 17.02 \\
			\hline
		\end{tabular}
		\label{tab:weight}
	\end{table}
	
\begin{figure}[th]
		\centering
		\includegraphics[width=\linewidth]{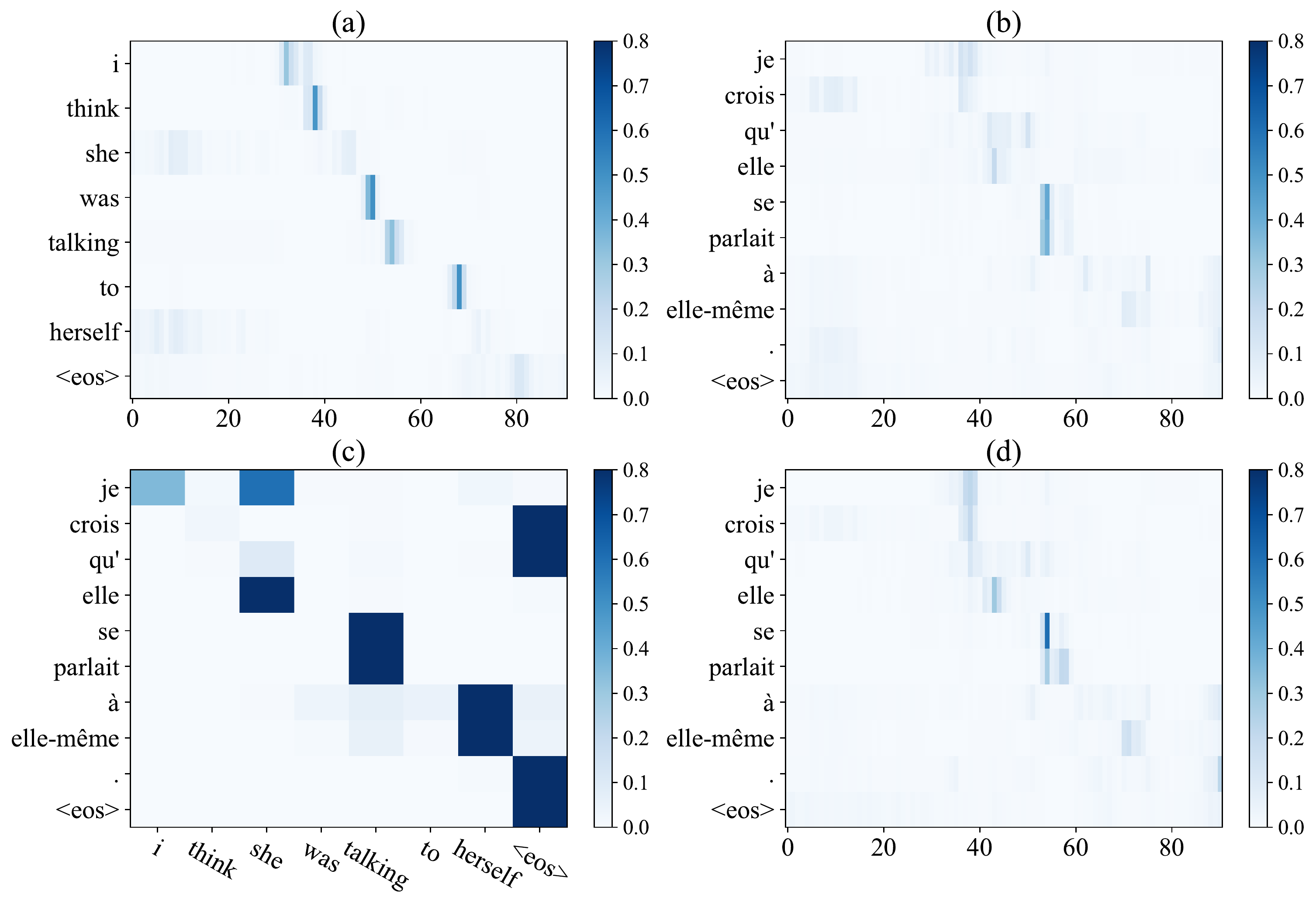}
		\caption{The visualizations of attention in different models. 
		$(a)$, $(b)$, $(c)$, $(d)$ are the encoder-decoder attention of ASR, end-to-end ST, MT and end-to-end ST with KD, respectively.}
		\label{fig:attention}
	\end{figure}

	We further analyze how knowledge from MT model helps ST through visualizations of the encoder-decoder attention. Figure \ref{fig:attention} shows an example. 
	The attentions of ASR (a) and MT (c) models have more confident than ST model. Each output token in the former two model concentrates on specific frames or tokens, especially for MT model, while the attention in ST (b) model tends to be smoothed out across many input frames. However, with the help of MT model, the attention of ST model with KD  (d)  becomes more concentrated. For example, the speech frames $l=45 \sim 55$ are corresponding to ``was talking'' in ASR (a), which can be translated to ``se parlait'' in French (c). The  attention in ST model with KD has more weights on frames $l=45 \sim 55$ than that in original ST model.
	
	\section{Conclusions}
	In this work, we present \textit{knowledge distillation} method to improve the end-to-end ST model by transferring the knowledge from MT model. Experiments on two language pairs demonstrate that with the instruction of MT model, end-to-end ST model can gain significant improvements. Although the end-to-end ST does not outperform pipeline system, it shows the potential to come close in performance. 
	In the future we will utilize other knowledges like the outputs from ASR model to further improve the performance of ST model.

	
	\bibliographystyle{IEEEtran}
	
	\bibliography{mybib}
\end{document}